\documentclass[twocolumn]{svjour3}          

\usepackage{graphicx}
\usepackage{amsfonts}
\usepackage[fleqn]{amsmath}

\usepackage{mathptmx}      
\usepackage{latexsym}
\journalname{Visual Computer}
\begin{document}

\title{Blind Motion Deblurring with Cycle Generative Adversarial Networks}

\author{Quan Yuan  \and Junxia Li \and Lingwei Zhang \and  Zhefu Wu\and Guangyu Liu }


\institute{Quan Yuan \at
          B-DAT, School of Automation, Nanjing University of Information Science and Technology, Nanjing, China \\
          \email{yuanquan.yuan@qq.com}           
          \and 
          Junxia Li \at 
          B-DAT, School of Automation, Nanjing University of Information Science and Technology, Nanjing, China \\
          \email{junxiali99@163.com}
          \and 
          Lingwei Zhang \at
          B-DAT, School of Automation, Nanjing University of Information Science and Technology, Nanjing, China \\
          \and
          Zhefu Wu \at
          B-DAT, School of Automation, Nanjing University of Information Science and Technology, Nanjing, China \\
          \and
          Guangyu Liu \at
          B-DAT, School of Automation, Nanjing University of Information Science and Technology, Nanjing, China \\
}

\date{Received: date / Accepted: date}

\maketitle

\begin{abstract}
	Blind motion deblurring is one of the most basic and challenging problems in image processing and computer vision. It aims to recover a sharp image from its blurred version knowing nothing about the blur process. Many existing methods use Maximum A Posteriori (MAP)  or Expectation Maximization (EM) frameworks to deal with this kind of problems, but they cannot handle well the figh frequency features of natural images. Most recently, deep neural networks have been emerging as a powerful tool for image deblurring. In this paper, we prove that encoder-decoder architecture gives better results for image deblurring tasks. In addition, we propose a novel end-to-end learning model which refines generative adversarial network by many novel training strategies so as to tackle the problem of deblurring. Experimental results show that our model can capture high frequency features well, and the results on benchmark dataset show that proposed model achieves the competitive performance.

\end{abstract}

\begin{figure}[htb]
	\centering
	\includegraphics[scale=0.49]{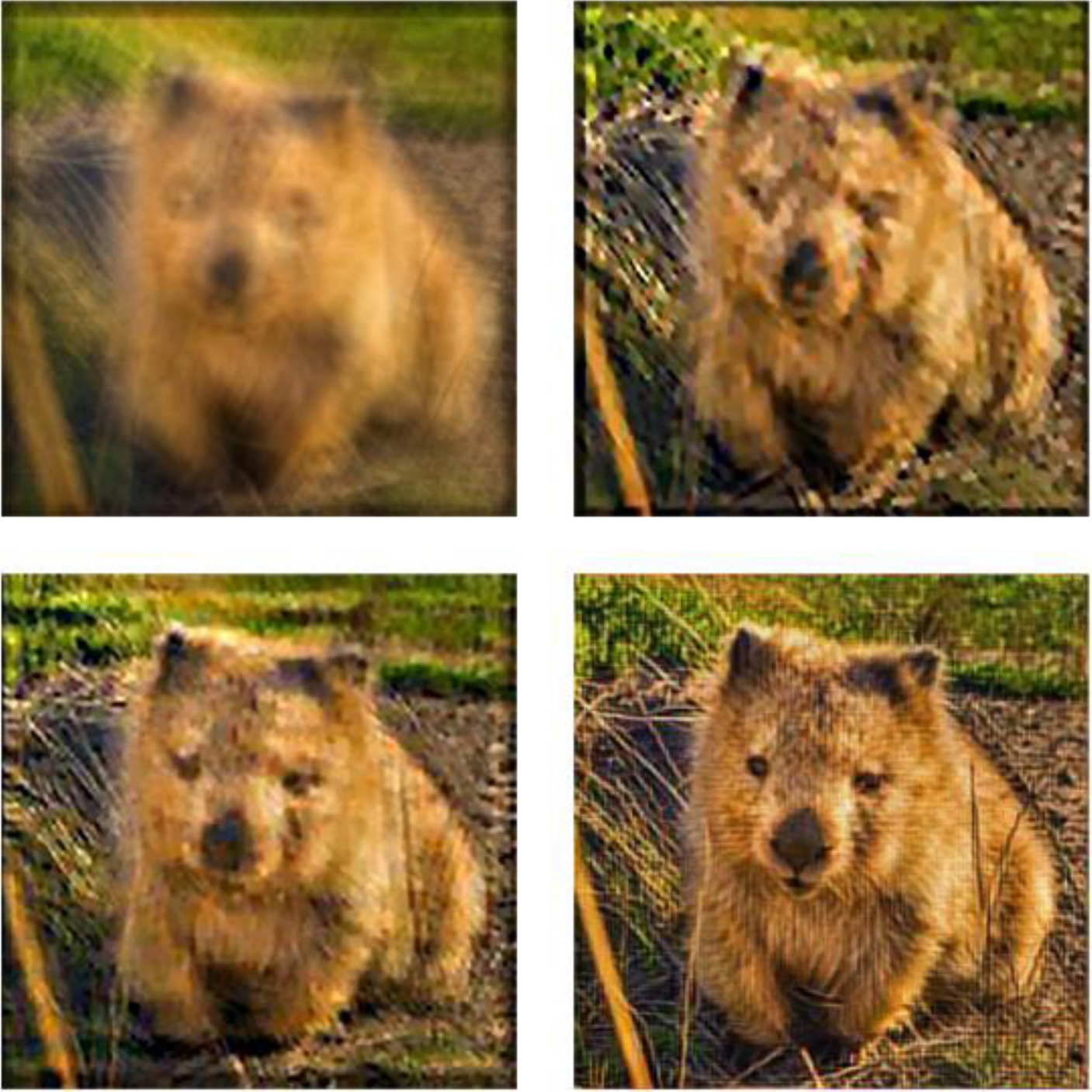}
	\caption{Deblurring results on a challenging motion blur image. From left to right : blurred image, the results by Pan et al.\cite{pan2016blind}, the results by Xu and Jia. \cite{xu2014deep} and ours. The results show that our model distinctly outperforms the competing methods.
		\label{fig:bear}}
\end{figure}

\section{Introduction}
\label{intro}

\begin{figure*}[htb]
	\centering
	\includegraphics[scale=0.23]{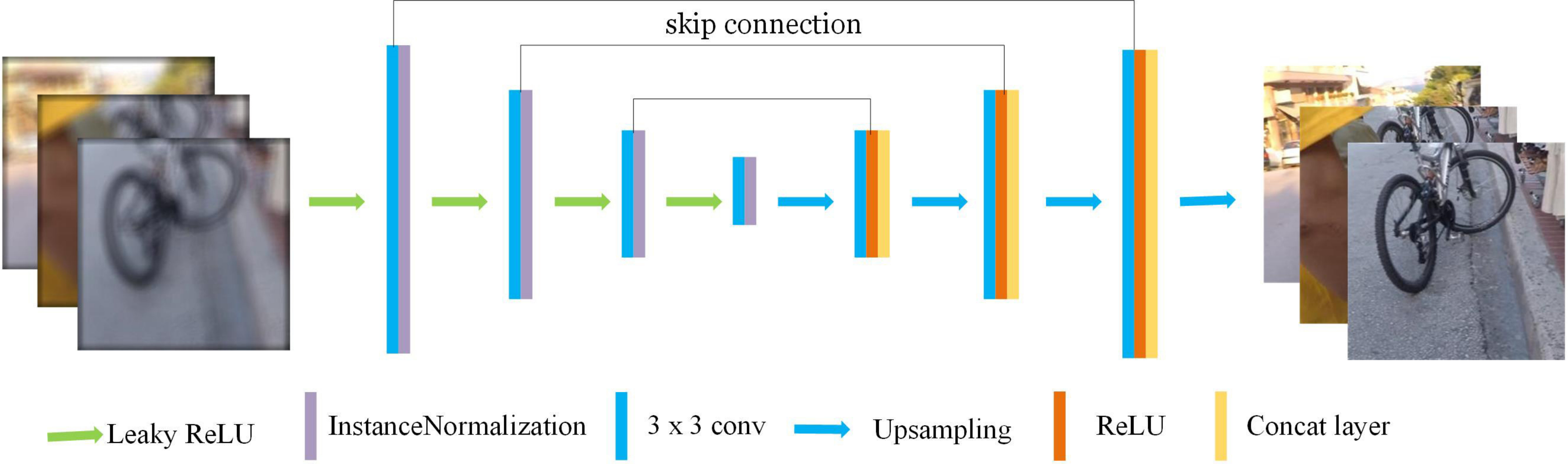}
	\caption{The generator network used in this work. The network contains encoder-decoder part with skip connection, Rectified Linear Unit and Instance Normalization. Equipped with novel techiques, our model can capture high frequency features, textures and details well, which outperforms in image deblurring tasks.}
\end{figure*}

	With the increase of digital cameras and mobile phones, a huge amount of high resolution images are taken every day \cite{khmag2018natural}\cite{fan2018detail}\cite{cambra2018generic}\cite{guan2018deep}\cite{yu2018deeper}, e.g. the latest Huawei Mate20 series mobile phones have over 60 megapixels. However, sensor shake is often inevitable, that resulting in undesirable motion blurring. Although  sharp images might be obtained by fixing devices or taking the images again, in many occasions, however, we have no chance to fix the devices or take the images again, for example in remote sensing\cite{ineichen2018high},	Video surveillance\cite{sun2018dppdl}, medical imaging\cite{hobbs2018physician} and some other related fields. Therefore, how to obtain sharp images from blurry images has been noticed by researchers in many fields for many years, but the problem still cannot be well solved due to the complexity of motion blur process and, most importantly, the rich details in high-resolution natural images. For example, whenever the blur kernel is complicated and the desired sharp images are rich in details, most existing methods may not produce satisfactory results, as we can see from Figure 1.

    Image deblurring problems are a kind of image degradation problems, which can be expressed as 
    
\begin{equation}
	I^{blur} = A(I^{sharp}) + n,
\end{equation}  
	where $I^{blur}$ is the given blurred image, $I^{sharp}$ is the sharp image, A is a degradation function and n denotes possible noise. In this work, we shall focus on the cases where the  degradation process is shift invariant, thereby the generation process of a blurred image is given by
\begin{equation}
	I^{blur} = I^{sharp} * k + n,
\end{equation}
    where * denotes 2D convolution and $k$ is the blur kernel. To obtain the sharp image and the blur kernel simultaneously, some commonly used approaches are MAP \cite{zhang2011sparse}\cite{chan1998total}, Variational Bayes \cite{fergus2006removing} \cite{levin2011efficient}. Lots of methods have been proposed and explored in the literature. For example, Chan and Wang \cite{chan1998total} proposed total variation to regularize the gradient of the sharp image. Zhang et al. \cite{zhang2011sparse} proposed a sparse coding method for sharp image recovering. Cai et al. \cite{cai2009blind} applied sparse representation to estimate sharp image and blur kernel at the same time. Although obtained moderately good results, these methods cannot apply to real applications and most importantly, cannot handle well high frenquency features. 
	
	To achieve fast image deblurring, it is straightforward to consider the idea of deep learning that pre-trains network models by a great deal of training data. Although the training process is computationally expensive, deep learning methods can process testing images very efficiently, as they only need to pass an image through the learnt network. Most existing deep based methods are built upon the well known Convolution Neural Network (CNN) \cite{pan2018physics} \cite{guo2018toward}. However, CNN tends to suppress the high frequency details in images. To relieve this issue, generative adversarial network (GAN) \cite{goodfellow2014generative}	is a promising idea. Kupyn et al. \cite{kupyn2017deblurgan} proposed a GAN based method that uses the ResBlocks architecture as the generator. Pan et al. \cite{pan2018physics} used GAN to extract intrinsic physical features in images.
    
    In this work, we prove that encoder-decoder network architecture performs better for image deblurring problems. Specifically, we propose a GAN based method using cycle consistency training strategy to ensure that the networks can model appropriately high frequency features. We build a cycle generator, which transfers the blurred images to the sharp domain and transfers the sharp images back to the blurred domain in a cycle manner. Different from previous works, we build two discriminators to distinguish the blurred and sharp images seperately. For generator, encoder-decoder based architecture performs better than ResBlock based architecture. Experimental results show that the method proposed in this work achieves the competitive performance.

\section{Related Works}

\begin{figure*}[htb]
	\centering
	\includegraphics[scale=0.27]{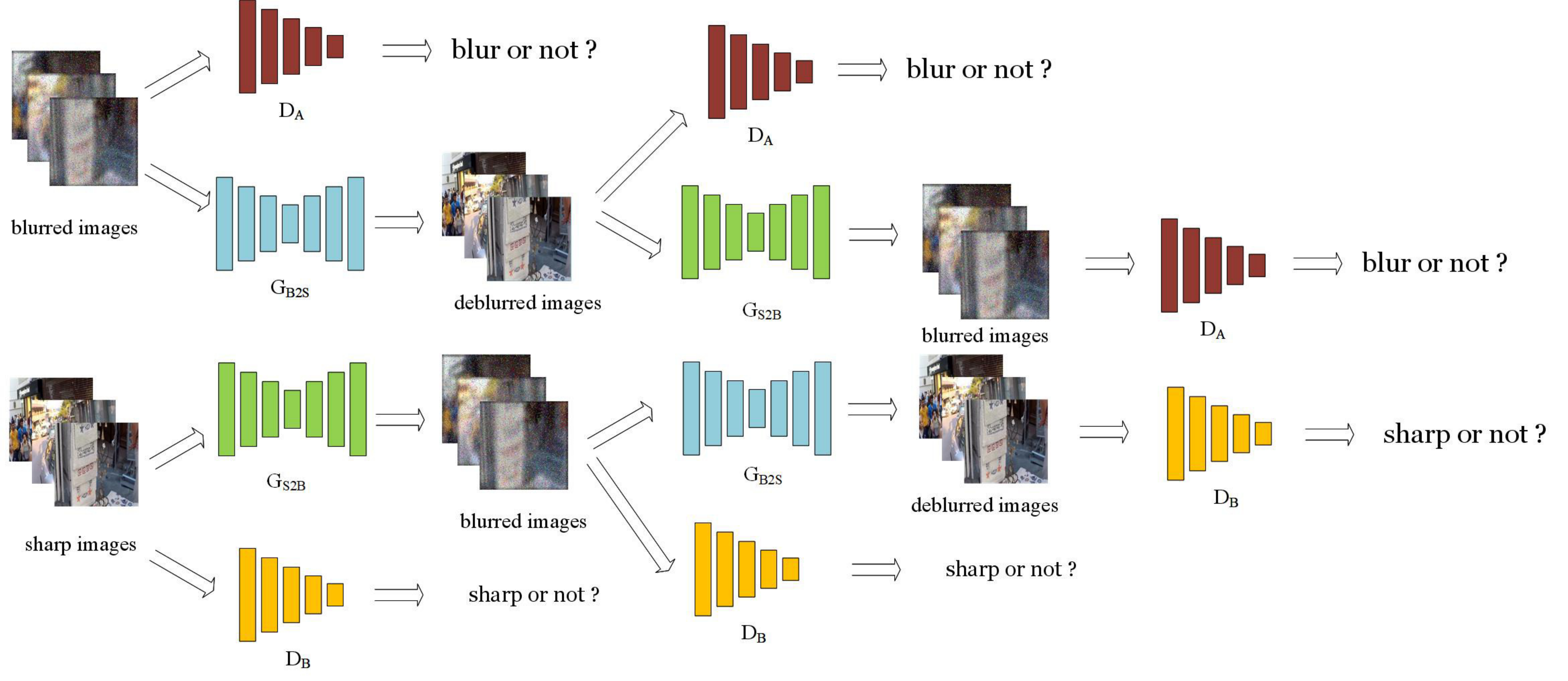}
	\caption{The network architecture with cycle consistency training strategy proposed in this work, including cycle generator $G_{B2S}$ and $G_{S2B}$ , which tries to generate sharp images given blurred image and generated blurred images given sharp images, respectively. And two discriminators which tries to distinguish the input images are blur or not, sharp or not, respectively. }
\end{figure*}

	Image deblurring is a classical problem in image processing and signal processing. We can divide image deblurring problems into learning-based methods and learning-free methods. 

	In learning-free methods, Most of works in this field suppose that deblurring is shift invariant and cause by motion \cite{liu2014blind}\cite{chandramouli2018plenoptic}\cite{kotera2018motion}, which can be treated as a deconvolution problem \cite{liu2014blind}\cite{krishnan2009fast}\cite{wang2018training}\cite{zhang2017learning}. 
	There are many ways to solve this, many works\cite{liu2014blind} used bayesian estimation to solve it, that is

\begin{equation}
	p(I^{sharp},k|I^{blur})\propto P(I^{blur}|I^{sharp},k)P(I^{sharp})P(k).
\end{equation}

	One commonly used deblurring method is based on the Maximum A MAP framework, where latent sharp image $I^{sharp}$ and blur kernel $k$ can be obtained by

\begin{equation}
\begin{aligned}
(k^*, I^{sharp*}) = \arg \max_{I^{sharp}, k} 
P(I^{blur}|I^{sharp}, k)P(I^{sharp})P(k).
\end{aligned}
\end{equation}

\begin{figure*}[htb]
	\centering
	\includegraphics[scale=0.34]{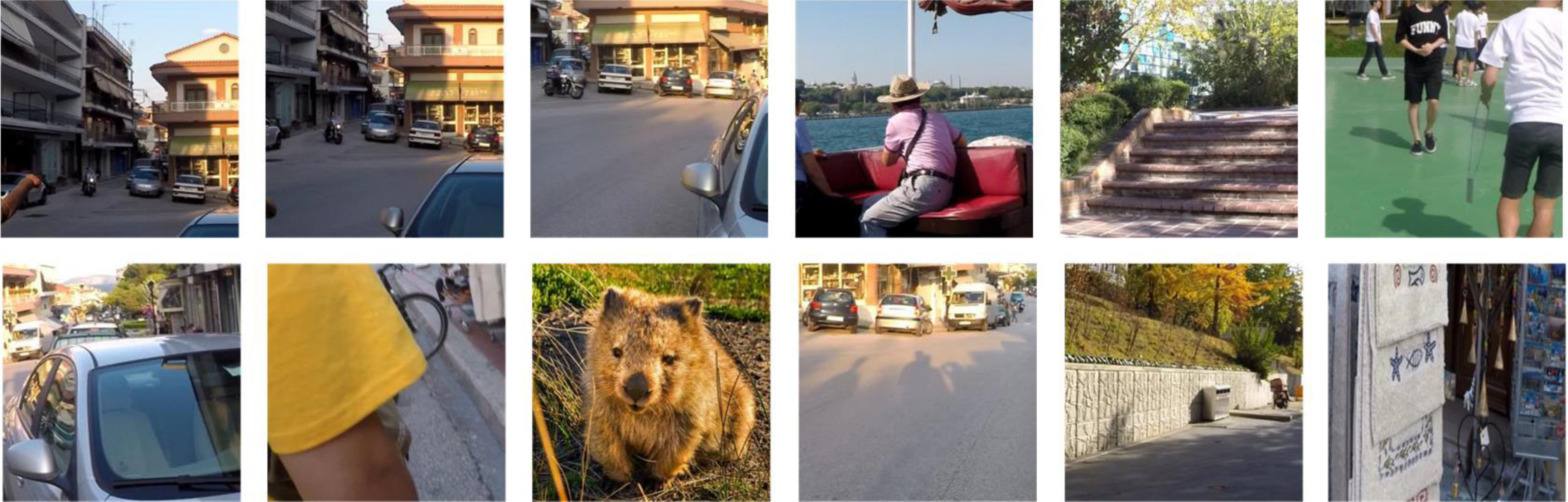}
	\caption{The dataset build in this work. These images are sampled from GoPRO dataset\cite{yeh2017semantic}. We tried as random as possible during sampling process to simulate the real images in different scenes.}
\end{figure*}

	Chan and Wang\cite{chan1998total} proposed a robust total variation minimization method which is effective for  regularizing the gradient or the edge of the sharp image. Zhang et al.\cite{zhang2011sparse} proposed a sparse coding method for sharp image recovering, which assumes that the natural image patch can be sparsely represented by a over-complete dictionary. Cai et al. \cite{cai2009blind}
	applied sparse representation for estimate sharp image and blur kernel at the same time. Krishnan et al.\cite{krishnan2011blind}
    found that the minimum of their loss function in many existing methods do not correspond to their real sharp images,so Krishnan et al. \cite{krishnan2011blind} proposed a normalized sparsity prior to handle this problems. Michaeli and Irani \cite{michaeli2014blind} found that multiscale properties can also be used for blind deblurring problems, so Michaeli and Iranli \cite{michaeli2014blind}proposed self-similarity as image prior.  Ren et al.\cite{ren2016image}proposed low rank prior for both images and their gradients. 

	Another commonly approach to estimate motion blur process is to maximize the marginal distribution
	
\begin{equation}
	\begin{aligned}
	p(k, I^{blur}) = \int p(k, I^{sharp}|I^{blur})dI^{sharp} \\
	               = \int p(I^{blur}|k)p(k)dI^{sharp}.
	\end{aligned}    
\end{equation}
	
	Fergus \cite{fergus2006removing} proposed a motion deblurring method based on Variational Bayes method. Levin \cite{levin2011efficient} proposed a Expectation Maximization (EM) method to estimate blur process. The above two approaches does have some drawbacks: it is hard to optimize, time consuming and cannot handle high frequency features well.

	While learning-based methods use deep learning techniques, which aims to find the intrinsic features which can be find by the models themselves through learning process. Deep learning \cite{lecun2015deep} has boost the research in related fields such as image recognition  \cite{krizhevsky2012imagenet}, image segmentation \cite{he2017mask} and so on. For deblurring problems using deep learning techniques, \cite{kupyn2017deblurgan} trained a CNN architecture to learn the mapping function from blurred images to sharp ones. \cite{pan2018physics} used a CNN architecture with physics-based image prior to learn the mapping function.

	One of the novel deep learning techiques is Generative Adversarial Networks, usually known as GANs,  introduced by Goodfellow \cite{goodfellow2014generative},  and inspired by the zero-sum game in game theory proposed by Nash  \cite{nash1951non} which has achieved many excited results in image inpainting \cite{yeh2017semantic}, style transfer \cite{isola2017image} \cite{zhu2017unpaired} \cite{johnson2016perceptual}, and it can even be used in other fields 	such as material science\cite{sanchez2018inverse}. The system includes a generator and a discriminator. Generator tries to capture the latent real data distribution, and output a new data sample, while discriminator tries to discriminate the input data is from real data distribution or not. Both the generator and the discriminator can build based on Convolutional Neural Nets \cite{lecun2015deep}, and trained based on the above ideas. Instead of input a random noise in origin generative adversarial nets \cite{goodfellow2014generative}, conditional GAN \cite{dai2017towards}
	input random noise with discrete labels or even images \cite{isola2017image} Zhu et al. \cite{zhu2017unpaired} take a step further, which based on conditional GAN and trained a cycle consistency objective, which gives more realistic images in image transfer tasks. Inspired by this idea, Isola \cite{isola2017image} proposed one of the first image deblurring algorithms based on Generative Adversarial Nets \cite{goodfellow2014generative}.

\section{Proposed Method}
	The goal of image deblurring model proposed in this work is to recover the sharp images given only the blurred images, with no information about the blur process, we build a generative adversarial network based model. A CNN was trained as a generator, given blurred images as inputs, outputs the sharp images. In addition, we also give a critic rules and train these models in an adversarial manner. We denote the two distributions as
	$I^{blur}\sim p_{data}(I^{blur}),I^{sharp}\sim p_{data}(I^{sharp})$   
	and two mapping functions, or so-called generator in GANs:
	$G_{B2S}: I^{blur} \rightarrow I^{sharp},$ $ G_{S2B}: I^{sharp} \rightarrow I^{blur} $.
	In addition, two discriminator $D_A$ and $D_B$ were introduced, $D_A$ tries to distinguish whether the input is blur or not while $D_B$ tries to distinguish whether the input is sharp or not. Our loss function contains two parts: adversarial loss and cycle loss. The architecture is shown in Figure 2, 3.

\subsection{Loss funtion}
	Our goal is to learn the mapping function between blurred domain B and sharp domain S given samples $\left\{ blur_i \right\}^M_{i=1]}$ where $blur_i\in B $ and  $\left\{ sharp_j \right\}^N_{j=1}$ where $ sharp_j\in S $.
	A combination of the following loss was used as our loss function:

\begin{equation}
	\mathcal{L} = \mathcal{L}_{adv} + \alpha \mathcal{L}_{cycle},
\end{equation} 
 
	\noindent where $\mathcal{L}, \mathcal{L}_{adv}, \mathcal{L}_{cycle}, \alpha$ are the total loss function, adversarial loss, cycle loss and their parameters, respectively. The adversarial loss tries to ensure the deblurred images as realistic as possible, cycle loss tries to ensure that the deblurred images can transfer back to the blur domain, which can also make the deblurred images as realistic as possible. 
	For the two mapping functions 
	$G_{S2B} : I^{sharp} \rightarrow I^{blur}, G_{B2S} : I^{blur} \rightarrow I^{sharp}$ which aims to transfer the sharp images to the blur domain and transfer the blurrred images to the sharp domain, respectively. The two corresponding discriminators $D_A, D_B$ tries to distinguish whether the input images are blur or not, sharp or not, respectively. The adversarial loss are as follows:
	The following loss function were proposed
\begin{equation}
	\mathcal{L}_{adv} = \mathcal{L}_{adv1} + \mathcal{L}_{adv2},
\end{equation}

	\noindent where
\begin{equation}
		\mathcal{L}_{adv1}   = \mathbb{E}_{I^{blur} \sim p_{data}(I^{blur})}[D_A(G_{B2S}(I^{blur}))],
\end{equation}

\begin{equation}
		\mathcal{L}_{adv2}   = \mathbb{E}_{I^{sharp} \sim p_{data}(I^{sharp})}[D_A(1 - G_{S2B}(I^{sharp}))],
\end{equation}

	\noindent where $D_A$ tries to distinguish whether the inputs are from target distribution or not, generators
	$G_{S2B}$ and $G_{B2S}$ tries to fool the discriminator and generate the images as realistic as possible.
	Isola et al.\cite{isola2017image} and Zhu et al.\cite{zhu2017unpaired} shown that least square loss\cite{mao2017least} can perform better than mean square loss in image style transfer tasks, Kupyn et al.\cite{kupyn2017deblurgan} 
	used least square loss \cite{mao2017least} for image deblurring tasks. So far, we don't know which loss objective performs better in image deblurring problems, mean square loss or least square loss\cite{mao2017least}, we have done some experiments to find out the better choice.

\begin{equation}
	\mathcal{L}_{cycle}  = \mathcal{L}_{cycle1}  + \mathcal{L}_{cycle2},
\end{equation}    
	where

\begin{equation}
\begin{aligned}
 & \mathcal{L}_{cycle1} = \frac{1}{N^{(i,j)} M^{(i, j)}} \\
 & \sum_{x=1}^{N^{(i,j)}} \sum_{y=1}^{M^{(i,j)}} (\sigma_{i,j}(I^{sharp})_{x,y}-\sigma_{i,j}(G_{B2S}(I^{blur}))_{x,y})^2,
\end{aligned}
\end{equation}
  
\begin{equation}
\begin{aligned}
& \mathcal{L}_{cycle2} = \frac{1}{N^{(i,j)} M^{(i, j)}} \\
& =\sum_{x=1}^{N^{(i,j)}}\sum_{y=1}^{M^{(i,j)}}(\sigma_{i,j}(I^{blur})_{x,y}-\sigma_{i,j}(G_{S2B}(I^{sharp}))_{x,y})^2,
\end{aligned}
\end{equation}

	\noindent where $\sigma_{i,j}$ is the feature map which obtained from the i-th maxpool layer after the j-th convolution layer from VGG-19 networks, and $N^{(i,j)}, M^{(i,j)}$ are the dimensions of the corresponding feature maps, the perceptual loss can capture high level intrinsic features which has been proved to work well in image deblurring\cite{kupyn2017deblurgan}, and some other image processing tasks\cite{isola2017image}\cite{zhu2017unpaired}.

	For cycle loss, which aims to make the reconstructed images and the input images as close as possible under some measurements, there are two classical choice for evaluation, L1 loss or mean square loss, Least Square loss\cite{mao2017least}or perceptual loss\cite{johnson2016perceptual}. The experiments shown that Perceptual loss\cite{johnson2016perceptual} can capture high frequency features in image deblurirng tasks, which gives more texture and details. So perceptual loss is used for evaluation in all experiments.

\subsection{Model Architecture}

	The goal of image deblurring problems is to map a low resolution inputs to a high resolution outputs. We use generative adversarial networks based model to deal with this problems. For discriminator, instead of classify the whole image is sharp or not,  we use PatchGAN based architecture tries to classify each image patch from the whole image, which gives better results in image deblurring problems. For the objective uses to distinguish whether the input is sharp or not, perceptual objective \cite{johnson2016perceptual} gives better results which not to evaluate the results in mean square or least square objectives, instead it tries to captures high freqency features of  the two given inputs. So PatchGAN based architecture\cite{isola2017image}\cite{zhu2017unpaired}	with perceptual loss\cite{johnson2016perceptual}were used as discriminator in all experiments. Experiments shown PatchGAN based architecture \cite{isola2017image}can achieves good results if the image patch is a quarter size of the input image, so in this work we choose patch $= 70 *70$ in all experiments according the input image size. For generator, many previous solutions for learning-based methods uses a encoder-decoder network such as Ronneberger et al. \cite{ronneberger2015u}as the generator, which shown in Figure 3, which tries to capture the high frequency features, it can distinguish blurred images from sharp images better than low-level features. Kupyn et al. \cite{kupyn2017deblurgan}used the architecture proposed by Johnson\cite{johnson2016perceptual} in deblurring problems as generator, which gives good performance.  Some comparative experiments shown in table 1 are given to find out which generator network architecture and  objective gives better results in image deblurring problems. The experiments shown that for image deblurring problems, the optimal choice for generator architecture is U-net based architecture and the optimal evaluation for optimization objective is least square loss. So the above generator architecture was used in our model in the following experiments. The whole model and the generator architecture were shown in Figure 2 and Figure 3.

\section{Experimental Results}

	We implement our model with Keras\cite{chollet2015keras} library. All the
	
	\begin{figure}[b]
		\centering
		\includegraphics[scale=0.48]{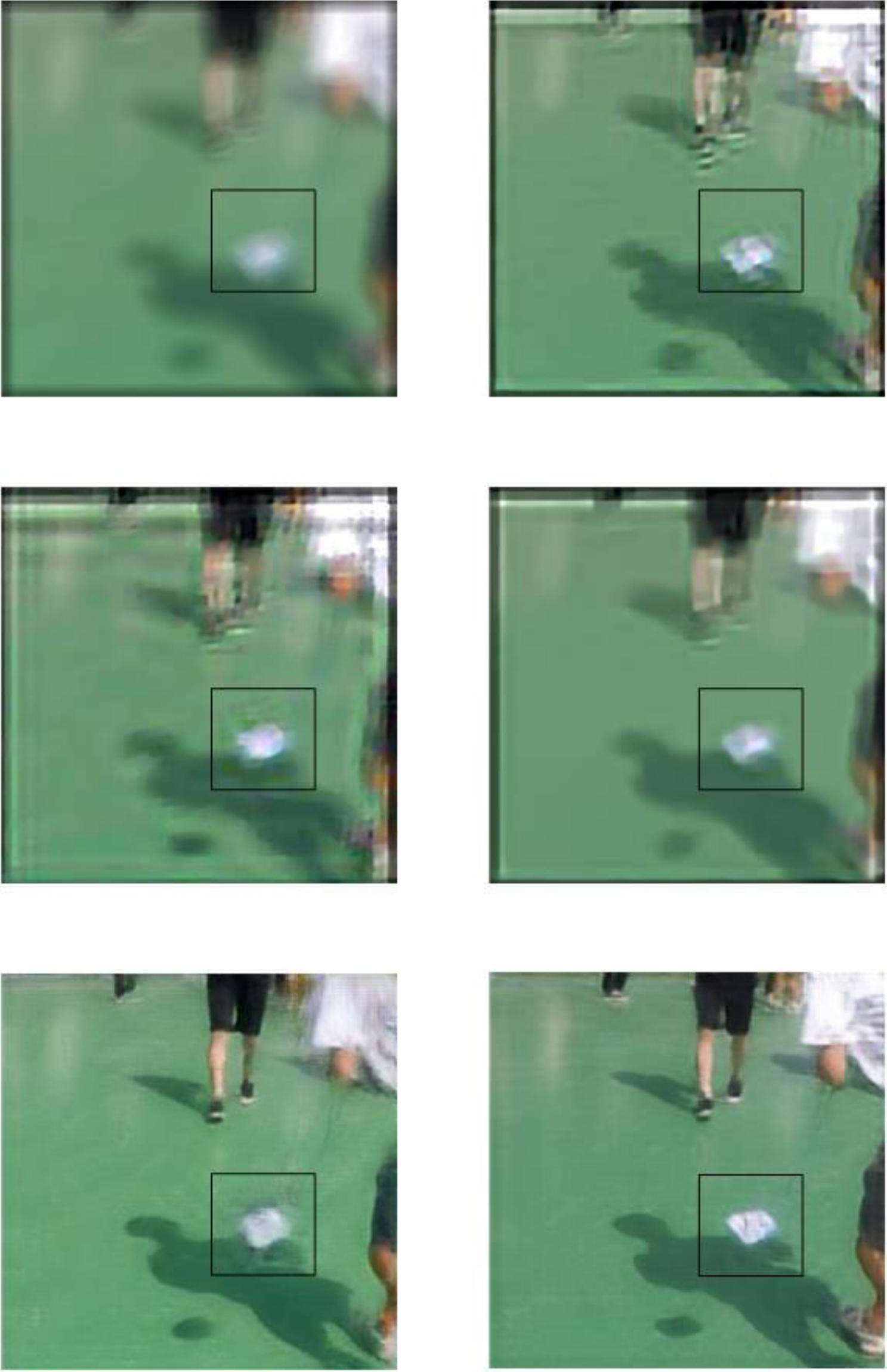}
		\caption{A comparison for deblurred images by our model and other algorithms in a  specific scene sampled from datasets. Top row : from left to right: blurred image, the results from Pan et al.\cite{pan2014deblurring}. Middle row: from left to right:  the results from Pan et al.\cite{pan2016blind}, the results from Xu and Jia.\cite{xu2014deep} . Bottom row:  from left to right: the results from Kupyn et al.\cite{kupyn2017deblurgan} and the results from Ours.}
	\end{figure}

\begin{table*}
	\textbf{Table 1}~~ The performance of different generator architecture dealing with deblurring problems. We evaluated on different full reference and non reference evaluations and the results shown that “U-net based + L2” architecture performed better than the other architectures.\\
	\setlength{\tabcolsep}{5.6mm}{
		\begin{tabular}{ccccccc}
			\hline 
			Methods & PSNR & SSIM & MS-SSIM & IFC & VIF & NIQE \\
			\hline
			Ours(Resblock based + L1)   & 18.1626 & 0.7386 & 0.8123 & 3.0014 & 0.8036 & 4.2355 \\
			Ours(Resblock based + L2) & 23.4854 & 0.8078 & 0.8808 & 3.8848 & 0.8554 & 5.0138\\
			Ours(U-net based + L1)      & 21.3191 & 0.7609 & 0.8800 & 3.1668 & 0.8401 & \textbf{5.6006}\\
			Ours(U-net based + L2) & \textbf{24.6043} & \textbf{0.8304} & \textbf{0.9029} & \textbf{4.2433} & \textbf{0.9067} & 5.2315\\
			\hline
	\end{tabular}}
\end{table*} 

\begin{table*}
	\textbf{Table 2}~~ A comparisive evaluations for our model and other algorithms for Figure \textbf{5} , which sampled a specific scene from the datasets\\ using different benchmark evaluations.The results shown that in this scene our model gives competitive results in most of the evaluations, \\and gives the results very close to the optimal in the rest evaluations\\
	\setlength{\tabcolsep}{6.5mm}{
		\begin{tabular}{ccccccc} 
			\hline
			Methods & PSNR & SSIM & MS-SSIM & IFC & VIF & NIQE \\
			\hline
			Pan et al.\cite{pan2014deblurring} &19.4989 & 0.8045 & 0.9756 & 0.2212 & 0.2515 & \textbf{29.4517} \\
			Pan et al.\cite{pan2016blind}    & 18.5531 & 0.7678 & 0.9744 & 0.1858 & 0.2946 & 28.1073\\
			Xu and Jia. \cite{xu2014deep}    &  19.2943 & 0.8004 & 0.9723 & 0.2858 & 0.4417 & 25.5722\\
			kupyn et al.\cite{kupyn2017deblurgan} & 25.9832 & 0.8912 & 0.9893 & 0.6692 & \textbf{0.6667} & 26.3751 \\ 
			Ours & \textbf{26.5584} & \textbf{0.9114} & \textbf{0.9934} & \textbf{0.7701} & 0.6580 & 28.8554\\
			\hline
	\end{tabular}}
\end{table*}

\begin{figure}
	\centering
	\includegraphics[scale=0.48]{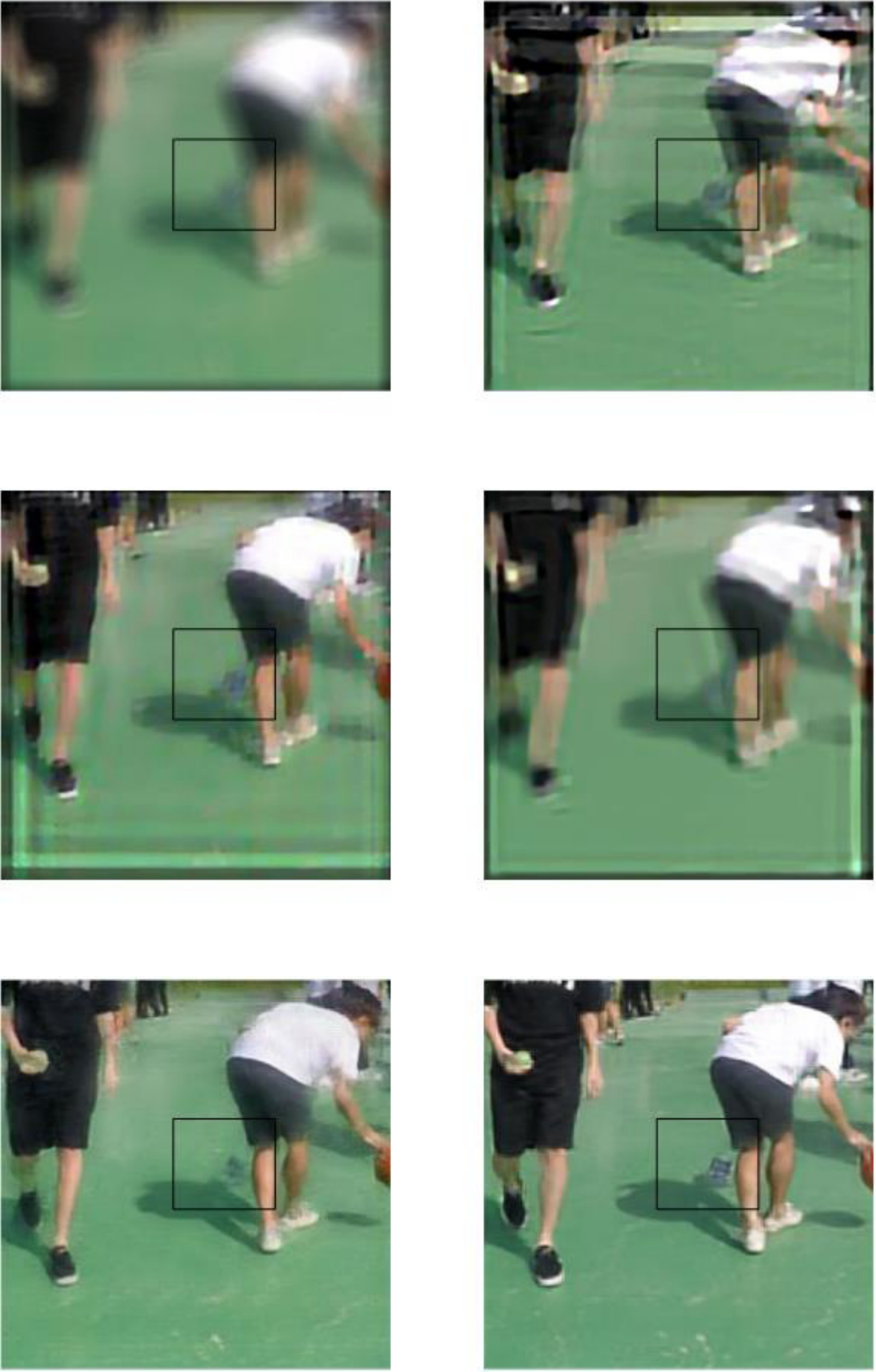}
	\caption{A comparison for deblurred images by our model and other algorithms in a  specific scene sampled from datasets. Top row : from left to right: blurred image, 
		the results from Pan et al.\cite{pan2014deblurring}. Middle row: from left to right:  the results from Pan et al.\cite{pan2016blind}, the results from Xu and Jia.\cite{xu2014deep} . Bottom row:  from left to right: the results from Kupyn et al.\cite{kupyn2017deblurgan} and the results from Ours.}
\end{figure}

\begin{figure}
	\centering
	\includegraphics[scale=0.48]{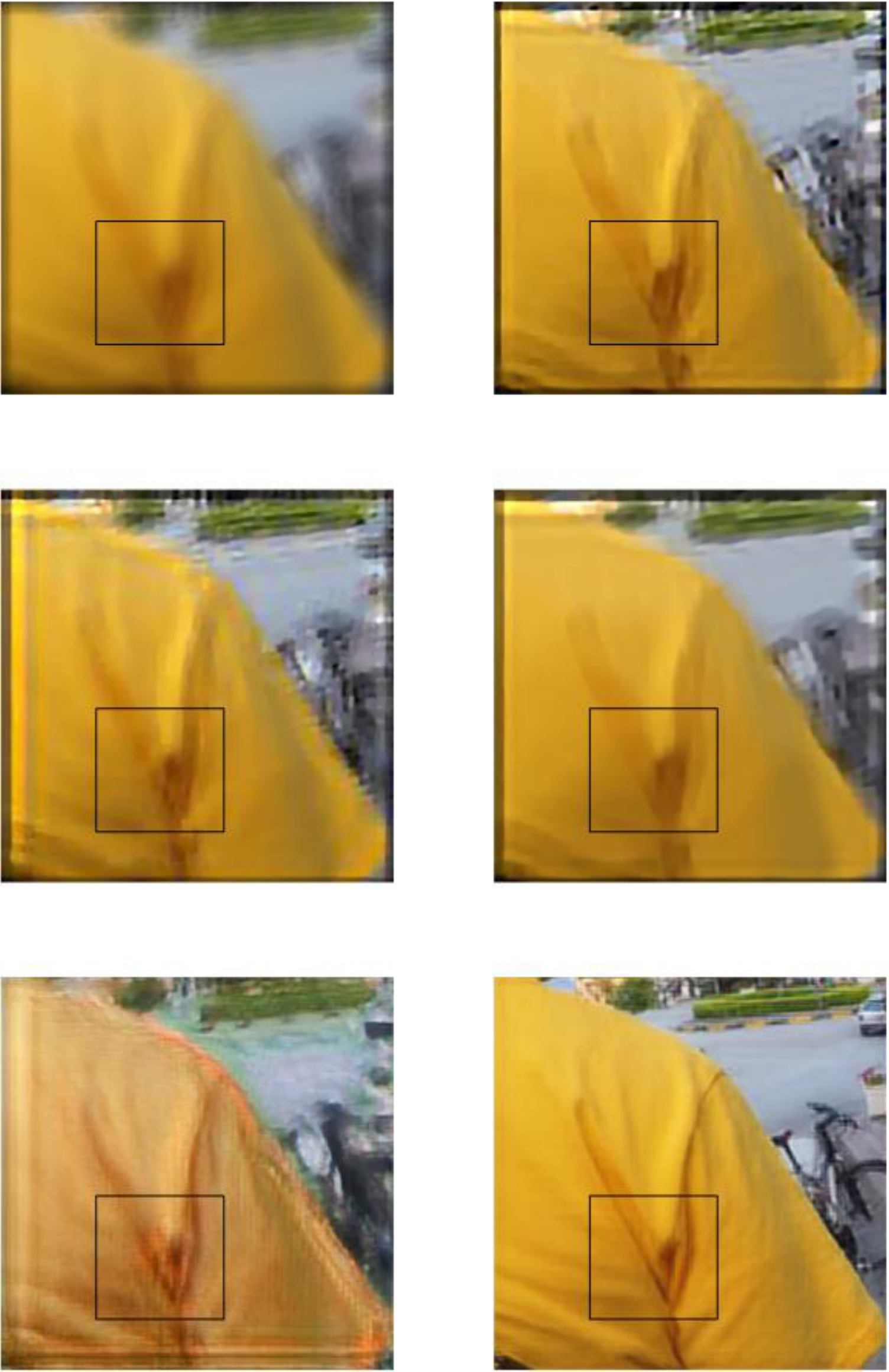}
	\caption{A comparison for deblurred images by our model and other algorithms in a  specific scene sampled from datasets. Top row : from left to right: blurred image, 
		the results from Pan et al.\cite{pan2014deblurring}. Middle row: from left to right:  the results from Pan et al.\cite{pan2016blind}, the results from Xu and Jia.\cite{xu2014deep} . Bottom row:  from left to right: the results from Kupyn et al.\cite{kupyn2017deblurgan} and the results from Ours.}
\end{figure}

	 \noindent experiments were performed on a workstation with NVIDIA Tesla K80 GPU. The network proposed was trained on the images sampled randomly from the GoPro datasets, and then divide into training sets and test sets. Figure 4 gives some images sampled from the datasets build in this paper. We sampled 10,100 images and resize each of them to $256 \times 256$
	 
	 \begin{table*}
	 	\textbf{Table 3}~~ A comparisive evaluations for our model and other algorithms for Figure \textbf{6} , which sampled a specific scene from the datasets\\ using different benchmark evaluations.The results shown that in this scene our model gives competitive results in most of the evaluations, \\and gives the results very close to the optimal in the rest evaluations \\
	 	\setlength{\tabcolsep}{5.5mm}{
	 		\begin{tabular}{ccccccc} 
	 			\hline
	 			Methods & PSNR & SSIM\cite{wang2004image} & MS-SSIM\cite{wang2003multiscale} & IFC\cite{sheikh2005information} & VIF\cite{sheikh2004image} & NIQE\cite{mittal2013making} \\
	 			\hline
	 			Pan et al.\cite{pan2014deblurring} & 18.6219 & 0.7220 & 0.9462 & 0.2380 & 0.3297 &\textbf{25.9329} \\
	 			Pan et al.\cite{pan2016blind}      & 18.6007 & 0.7059 & 0.9272 & 0.3967 & 0.5401 & 25.2934\\
	 			Xu and Jia. \cite{xu2014deep}      & 21.0771 & 0.7997 & 0.9562 & 0.6519 & 0.5382 & 25.2934\\
	 			kupyn et al.\cite{kupyn2017deblurgan} & 24.7594 & 0.8716 & \textbf{0.9714} & 1.0319 & 0.6369 & 23.5593\\
	 			Ours & \textbf{27.1089} & \textbf{0.9003} & 0.9700 & \textbf{1.5143} & \textbf{0.8977} & 25.4217\\
	 			\hline
	 	\end{tabular}}
	 \end{table*}
 
 \begin{table*}
 	\textbf{Table 5}~~ A comparisive evaluations for our model and other algorithms for Figure \textbf{8} , which sampled a specific scene from the datasets\\ using different benchmark evaluations.The results shown that in this scene our model gives competitive results in most of the evaluations, \\and gives the results very close to the optimal in the rest evaluations\\
 	\setlength{\tabcolsep}{5.5mm}{
 		\begin{tabular}{ccccccc} 
 			\hline    
 			Methods & PSNR & SSIM\cite{wang2004image} & MS-SSIM\cite{wang2003multiscale} & IFC\cite{sheikh2005information} & VIF\cite{sheikh2004image} & NIQE\cite{mittal2013making} \\
 			\hline
 			Pan et al.\cite{pan2014deblurring} & 17.2663 & 0.5999 & 0.9644 & 0.2381 & 0.1564 & 20.2053\\
 			Pan et al.\cite{pan2016blind}      & 16.1910 & 0.5613 & 0.9556 & 0.3025 & 0.2213 & 21.7997\\
 			Xu and Jia. \cite{xu2014deep}      & 17.2508 & 0.5614 & 0.9651 & 0.4181 & 0.2580 & \textbf{22.4268}\\
 			kupyn et al.\cite{kupyn2017deblurgan} & 13.1351 & 0.6728 & 0.9574 & 0.9368 & 0.3311 & 19.2725\\
 			Ours & \textbf{19.4527} & \textbf{0.6872} & \textbf{0.9825} &\textbf{0.9454} & \textbf{0.3805} & 21.0580\\
 			\hline
 	\end{tabular}}
 \end{table*}

\begin{figure}
	\centering
	\includegraphics[scale=0.48]{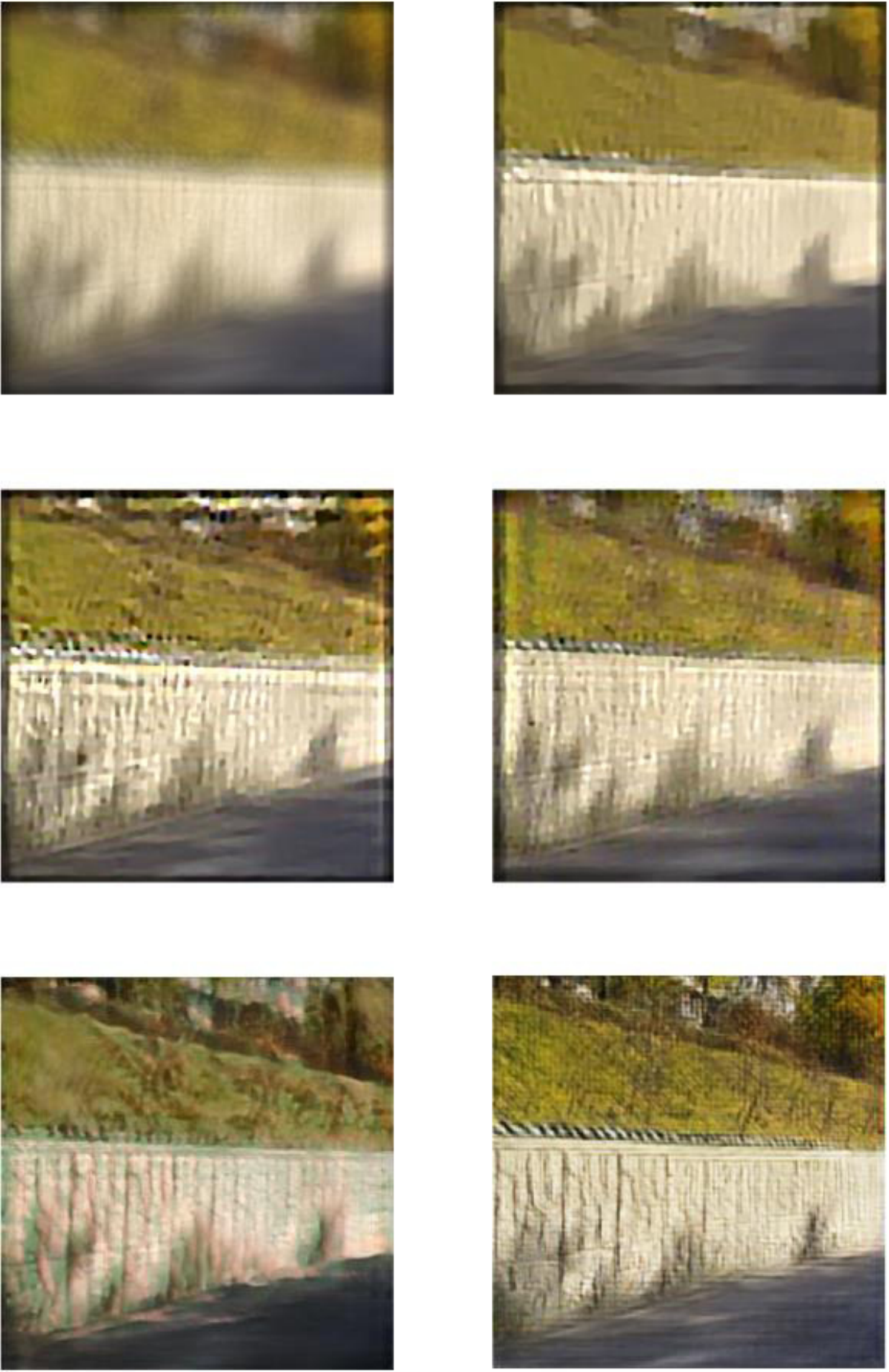}
	\caption{A comparison for deblurred images by our model and other algorithms in a  specific scene sampled from datasets. Top row : from left to right: blurred image, 
		the results from Pan et al.\cite{pan2014deblurring}. Middle row: from left to right:  the results from Pan et al.\cite{pan2016blind}, the results from Xu and Jia.\cite{xu2014deep} . Bottom row:  from left to right: the results from Kupyn et al.\cite{kupyn2017deblurgan} and the results from Ours.}
\end{figure}

	 \noindent $\times 3$, applied blur process proposed by Kupyn et al.\cite{kupyn2017deblurgan}. We randomly choose 10,000 images for training precedure and the rest 100 images for testing. We train the network with a batch size of 2, giving 50 epochs over the training data. The reconstructed images are regularized with cycle consistency objective with a strength of 10. we use paired data to train the model in 50 epochs. No dropout technique was used since the model does not overfit within 50 epochs. For the image pairs used during training and testing process, we use the method similar to the one proposed by Kupyn et al.\cite{kupyn2017deblurgan}, which produce more realistic motion blur than other existing methods \cite{xu2014deep} \cite{sun2015learning}. For the  optimization procedure, we perform 10  steps on $D_A$ and $D_B$, and then one step to $G_{S2B}$ and $G_{B2S}$. We use Adam \cite{kingma2014adam} optimizer with a learning rate of $2 \times 10^{-3}$ in the first 40 epochs, and then linearly decay the learning rate to zero in the following epochs to ensure the convergence. Both generator and discriminators were applied with Instance Normalization to boost the convergence. We choose some deblurred results from  different scenes, which are given in Figure 5, 6, 7, 8 and Table 2, 3, 4, 5, respectively. For image evaluation, most works use full reference measurments PSNR and SSIM in all their experiments \cite{pan2014deblurring} \cite{pan2016blind} \cite{pan2018deblurring} \cite{pan2018physics} \cite{wang2004image}, which need reference images (groundtruth) during assessments. For other image assessments, VIF \cite{sheikh2004image} captures wavelets features which focus on high frequency features, IFC \cite{sheikh2005information} puts more wights on edge features. Lai et al \cite{lai2016comparative} points out that the full reference image assessments VIF and IFC is better than PSNR and SSIM. So in this paper,  we take a step further, which use PSNR, SSIM\cite{wang2004image} and some new full reference methods MS-SSIM \cite{wang2003multiscale}, IFC\cite{sheikh2005information},  VIF\cite{sheikh2004image} and one No Reference image quality assessment NIQE \cite{mittal2013making} in all experiments.
     For the experimental comparision, we choose different learning-free methods proposed by Pan et al.\cite{pan2014deblurring}, Pan et al.\cite{pan2016blind} and Xu and Jia.\cite{xu2014deep}, and for fairness, we also choose one learning-based method proposed by Kupyn et al.\cite{kupyn2017deblurgan} for comparisions. All the salient regions were pointed out in each images.

\begin{table*}
	\textbf{Table 4}~~ A comparisive evaluations for our model and other algorithms for Figure \textbf{7} , which sampled a specific scene from the datasets\\ using different benchmark evaluations.The results shown that in this scene our model gives competitive results in most of the evaluations, \\and gives the results very close to the optimal in the rest evaluations.\\
	\setlength{\tabcolsep}{5.5mm}{
		\begin{tabular}{ccccccc} 
			\hline    
			Methods & PSNR & SSIM\cite{wang2004image} & MS-SSIM\cite{wang2003multiscale} & IFC\cite{sheikh2005information} & VIF\cite{sheikh2004image} & NIQE\cite{mittal2013making} \\
			\hline
			Pan et al.\cite{pan2014deblurring} & 19.2479 & 0.8202 & 0.9724 & 0.2475 & 0.2161 & 28.0291 \\
			Pan et al.\cite{pan2016blind}      & 18.5358 & 0.8030 & 0.9593 & 0.2297 & 0.2625 & \textbf{28.9056}\\
			Xu and Jia. \cite{xu2014deep}      & 19.4828 & 0.8259 & 0.9716 & 0.3699 & 0.3334 & 27.1181\\
			kupyn et al.\cite{kupyn2017deblurgan}&  19.6869 & 0.8173 & 0.7415 & 0.5814 & 0.4538 & 23.0904\\
			Ours & \textbf{25.2607} & \textbf{0.9308} & \textbf{0.9813} & \textbf{1.3580} & \textbf{0.6851} & 25.9405\\
			\hline
	\end{tabular}}
\end{table*}
     
     The results shown that the model proposed in this work outperformed many existing image deblurring models, it can recover more high frequency textures and details, the salient regions were pointed out in Figure 5, 6, 8. Our model outperfiorms many existing learning-free and learning-based methods in most full reference assessments and human visualization evaluations, the results are shown in Table 2, 3, 4, 5. But our model does not perform well in no reference assessments. Some results get higher score (e.g. in Table 2, 3, 4, 5) does not perform well in human visualization evaluations, so we think that NIQE may not applicable for image deblurring problems assessments. It also shown that our models can handle blur caused by motion or camera shake, the recovered image has less artifacts comparing to many existing methods.

\section{Conclusions}

    We shown that encoder-decoder based architecture performs better for image deblurring problems comparing the Resblock based architecture. For optimization objectives, least square loss performs better than mean square loss. The experiments shown that the model proposed in this work can deal with image deblurring problems well  without giving any domain specific knowledege. It can recover more high frequency textures and details, which not only outperform many competitive methods in many different full reference and no reference image quality assessments but also in human visualization evaluation. It also shown that our models can handle blur caused by motion or camera shake, the recovered image has less artifacts comparing to many existing methods.

\section{Compliance with Ethical Standards}
	\textbf{Fundings}: This study was funded by the National Natural Science Foundation of China (NSFC) under grant 61502238, grant 61622305, grant 6170227 and the Natural Science Foundation of Jiangsu Province of China (NSFJS) under grant BK20160040.

	\noindent \textbf{Conflict of Interest}: The authors declare that they have no conflict of interest.

\begin{acknowledgements}
	We thank Guangcan Liu, Yubao Sun, Jinshan Pan and Jiwei Chen for their helpful discussions and advices.
\end{acknowledgements}

\bibliographystyle{spmpsci}      
\bibliography{citation}   
	
\end{document}